# A Compact Gated-Synapse Model for Neuromorphic Circuits

Alexander Jones and Rashmi Jha

*Abstract*—This work reports a compact behavioral model for gated-synaptic memory. The model is developed in Verilog-A for easy integration into computer-aided design of neuromorphic circuits using emerging memory. The model encompasses various forms of gated synapses within a single framework and is not restricted to only a single type. The behavioral theory of the model is described in detail along with a full list of the default parameter settings. The model includes parameters such as a device's ideal set time, threshold voltage, general evolution of the conductance with respect to time, decay of the device's state, etc. Finally, the model's validity is shown via extensive simulation and fitting to experimentally reported data on published gated-synapses.

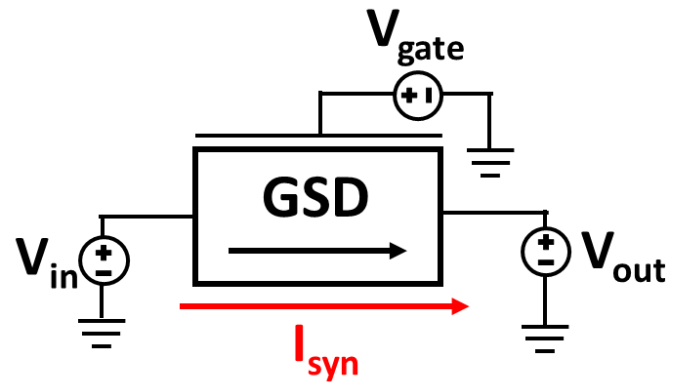

Figure 1. Diagram of a gated-synaptic device (GSD). The device is normally programmed via the gate terminal, while $v_{in}$ and $v_{out}$ act as pre-synaptic and post-synaptic analysis terminals, respectively.

## I. INTRODUCTION

DEEP learning has re-kindled interest in artificial neural networks in recent years [1]-[4]. Neuromorphic computing, the hardware implementation of neural networks, has also become an important research topic [5]-[8]. In the past decade a popular method of implementing synapses within neuromorphic hardware has been via two-terminal memristive or hysteretic devices [9]-[11]. More recent research shows a growing interest in gated-synaptic devices (GSDs) due to its capabilities not available in two-terminal devices such as simultaneous read/write [12]-[18]. A typical GSD usually possesses input/output terminals along with some form of gate as shown in Fig. 1. A programming bias is applied to the gate to control the conductance between input and output terminals.

Despite GSDs becoming an increasingly popular topic of research in recent years, no generic behavioral model exists yet to describe them in a SPICE/Verilog-A compatible framework for computer-aided design and simulation. The lack of such a model prohibits a broader community of circuit designers to implement GSDs into their designs. By developing a generic model that can encapsulate defining behaviors such as input/output conductance configurability via a gate bias, non-linear temporal conductance behavior, and dynamic state retention and volatility will allow for more accurate transient analysis of GSDs and the networks they inhabit.

Previous studies have shown how the transient dynamics of neural networks can affect a network's accuracy and learning capability [19]-[22]. Work done by Tyasnurita et al. demonstrated the use of time delay within a neural network to extract hyper-heuristic features from data for open vehicle routing [19]. Some have developed and verified (against MNIST) complete tools to simulate neuromorphic architectures within the transient domain such as Shahsavari and Boulet's N2S3 framework [20]. Other work within the time dynamics of neural networks has focused on topics such as noise vs. information. Boerlin and Denève approached this topic by studying the time-dynamic interplay between synapses and neurons [21]. Others have studied the realm of biological systems such as Abbott and Regehr whose work focused on how the timing of pre- and post-synaptic signals to a synapse affected the potentiation of the synapse [22].

This continual study of transient properties within neural networks and neuromorphic architectures demonstrates the need for a behavioral model for a GSD that can handle the transient domain. If a generic device model can be developed to capture the various behaviors of previously published GSDs [12]-[18] in addition to potential behavior of future GSDs, it could assist in the development of new and exciting neural networks and neuromorphic architectures that rely on time dynamics for learning. The following work will propose such a model for GSDs that emulates both previously seen and potential future device behavior and dynamics.

This work was initially submitted for review on Apr. 28, 2020. This work was supported by the National Science Foundation under award number CCF-1718428.

A. Jones is with the University of Cincinnati, Cincinnati, OH 45220 USA. (e-mail: jones2a5@mail.uc.edu).

R. Jha is with the Electrical Engineering and Computer Science Dept. at the University of Cincinnati, Cincinnati, OH 45220 USA (e-mail: jhari@ucmail.uc.edu).

## II. Device Features and Behavior

In order to develop a behavioral GSD model, published work done on experimentally proven devices that qualify as GSDs should be studied in order to observe critical behavioral characteristics that exist within each device. In order to be classified as a GSD, the device must possess the following:

- A primary device channel between two terminals that can qualify as input and output nodes for current analysis.
- A gate terminal which can be used to either potentiate or depress the device.
- Some degree of non-volatile behavior when programming the device via the gate terminal.

Over the past few years, multiple publications have been made on various designs of devices that would qualify as GSDs. In order to create the behavioral model for this work, seven of these recently published devices were studied [12]-[18].

The devices that appear in [12]-[18] vary greatly in design and core operational mechanisms. This device spectrum varies from highly CMOS-compatible device designs such as Lim et al. [15] to Bao et al.'s device that relies on liquid electrolytes for its device channel [12]. Two of the devices studied rely on a pseudo-gate to potentiate/depress the device in the form of light being applied to one end of a two-terminal device from Murdoch et al. [16] and Tan et al. [17]. These devices still qualify as GSDs since the applied light programs the device whilst voltage applied to the same terminal has either a lesser or no effect. The remaining devices studied for this model lie in the regime of more traditional oxide-based devices such as resistive RAM, but with a gate added to the device's design [13], [14], [18].

Within the wide range of GSDs studied for this model many different behaviors can be seen, but with a core of universal traits. If one studies all seven previously published devices, there is a set of properties that appear to within each one.

- A "general shape" of a conductance curve when being potentiated.
- An amount of time is required for the device to reach its highest conductive state.
- Minimum and maximum conductance values.
- Some sort of state-based short-term decay when no potential is being applied to the gate of the GSD.
- Some degree of long-term plasticity.

Among the universal properties observed across all devices, there are certain properties that appear in one or more of the devices, but not all. For example, certain devices such as Bao et al. [12] and Herrmann et al. [14] possess gate threshold voltages where the device is neither potentiated (or depressed) if the gate voltage exceeds the threshold voltage value. Other devices are more sensitive to negative gate voltage that can more quickly depress the device [15], or some can simply be reset via assistance from device channel bias [16], [17]. Devices can also possess diode-like behavior when the GSD's I/O channel is placed into reverse bias [15], [16]. Finally, the device shown in [13] has a unique property where the redox reaction required to potentiate/depress the device requires negative voltage to potentiate and positive voltage to depress (instead of vice versa). Despite this final property only being visible in a single device, the nature of the property is critical to the device's operation, and therefore should be considered as an option within the behavioral model.

In order to encapsulate all the features observed in the devices studied in this work, the behavioral GSD model includes an array of 14 user-defined parameters that describe each behavioral feature in some way and can be seen in Table I. Each of these terms is described in greater detail in Section

TABLE I
USER-DEFINED MODEL PARAMETERS

| Parameter | Default Value | Range | Description |
|---|---|---|---|
| $g_c$ | 0.0 | 0 to 1 | Central fitting parameter that controls the device's overall transient conductance curve shape. |
| $b_{rev}$ | 1.0 | 0 to 1 | Defines whether the behavior during reverse bias is that of a memristor/resistor ($b_{rev}$=1) or diode ($b_{rev}$=0), or in between (0<$b_{rev}$<1). |
| $g_{min}$ | 1e-11 S | <$g_{max}$ | Minimum conductance value. |
| $g_{max}$ | 1e-6 S | >$g_{min}$ | Maximum conductance value. Acts as an asymptote for inverse exponential/sigmoid portion of conductance equation. |
| $t_{set}$ | 1e-6 s | >0 | Ideal set time of device (assuming no decay and potentiation @1V above threshold voltage on $v_{gate}$). |
| $v_t$ | 0.0V | ≥0 | Threshold voltage for potentiation/depression. |
| $n_{amp}$ | 1 | >0 | Controls amplified depression when negative bias is applied to $v_{gate}$. The higher the value of $n_{amp}$, the higher the amplification. |
| $o_c$ | 0.0 | 0 to 1 | Dictates control of how much the bias across the device channel ($v_{in}$-$v_{out}$) influences the effective voltage applied to the device via $v_{gate}$. $o_c$=0 means it has no influence, and $o_c$=1 means the channel bias is fully included in calculating $V_{eff}$. |
| $t_c$ | 0.0 | 0 to 1 | Dictates control of how much $v_t$ emphasis occurs when calculating the change in $x$ and $x_{min}$. $t_c$=0 is none, while $t_c$=1 is a perfect difference between $v_{gate}$ and $v_t$. |
| $r_{stp}$ | 0.0 | ≥0 | Rate of decay for short-term plasticity. Magnitude often inversely proportional to set time ($t_{set}$). |
| $q_{ltp}$ | 0.0 | 0 to 1 | Controls the quality of the long-term potentiation rate of the device. The higher the value of $q_{ltp}$, the closer the long-term potentiation conductance value is to $g_{max}$ as the device is programmed. |
| $r_{ltp}$ | 0.0 | ≥0 | Controls the rate of decay for long-term plasticity. The higher the value, the sharper the rate of decay for the long-term plasticity state. |
| $f$ | 1 | 1, -1 | Indicates if the polarity at which the device is potentiated/depressed is flipped or not. $f$=1 means it's not flipped, while $f$=-1 means it is. |
| $x_{start}$ | 0.0 | 0 to 1 | Parameter that defines the initial conductance state of the device. $x_{start}$=0 means the device is in its lowest conductance state, and $x_{start}$=1 means it's in its highest conductance state. |

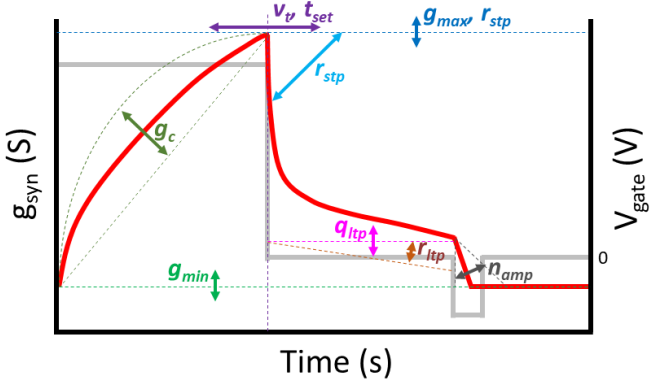

Figure 2. Diagram of a typical potentiation/decay/reset curve for a GSD. The conductance of the synapse increases with time when being potentiated, with the state beginning to quickly decay once potentiation ceases. The device can be fully reset with a negative gate bias if desired. Many of the user-defined parameters within the model can visibly manipulate the shape of this curve as shown in the diagram.

III. The only term that captures a behavior not mentioned previously is the term, $x_{start}$. This final term can be used to help define a GSD in a pre-potentiated or virgin conductance state. This parameter can also assist neural network designers looking to utilize the GSD model in initializing an array of devices with random initial conductance values to obtain initially random weights within their network.

From a conceptual perspective, if a GSD is potentiated via gate bias for a period of time, then left to have its state decay to some intermediate conductive state, and then finally has its state reset by a bias of opposite polarity, each device would experience a curve similar the one shown in the diagram in Fig. 2 (when in forward bias). Several parameters from Table I would have a degree of influence on this curve, which can be seen via the labels in Fig. 2. Other parameters such as $o_c$, $t_c$, $f$, and $x_{start}$ would influence this curve in more nuanced fashion. Parameters such as $b_{rev}$ would hold influence in a situation where the device is in reverse bias.

## III. MODEL DESCRIPTION

In order to utilize the user-defined parameters described in Table I, the model uses a series of equations in order to encapsulate all the behavior seen in studied devices.

### A. Device Current and Conductance

The current ($I_{syn}$) through the GSD is defined by

$$I_{syn} = \begin{cases} \Delta V \geq 0, g_{syn}\Delta V \\ \Delta V < 0, b_{rev}g_{syn}\Delta V + (1 - b_{rev})g_{syn}(e^{\Delta V} - 1) \end{cases} \quad (1),$$

where $g_{syn}$ is the conductance of the device and $\Delta V$ is the voltage difference across the device channel given by

$$\Delta V = v_{in} - v_{out} \quad (2).$$

The $b_{rev}$ term within (1) varies between 0 and 1 and dictates how the device behaves during reverse bias (as described in Section II). $b_{rev}$ is the primary reason why the equation is a piecewise function with respect to $\Delta V$. When under reverse bias ($\Delta V<0$) and when $b_{rev}<1$, a pseudo-diode equation starts to take control of the device's current (fully taking control when $b_{rev}=0$). This part of the equation is kept simple in order to still embody the behavioral nature of the model. This diode portion of the equation is limited by the device's conductance ($g_{syn}$), as the only device that shows a wide range of reverse bias behavior (Lim et al. [15]) shows the device in reverse bias still increasing in conductance as it is being programmed. Other devices might exhibit diode behavior in reverse bias that does not change as the device is programmed and might rely on other limiting terms (e.g. $g_{min}$). The devices studied here do not possess such behavior, however.

The reverse bias nature of an unprogrammed GSD can be seen in Fig. 3. As $b_{rev}$ is varied from 0 to 1 within the Verilog-A module, the IV curve changes from a linear memristive curve to a curve that resembles a diode in reverse bias. The forward bias component of the IV curve is fully linear as it does not include the diode component due to the diode behavior not being the primary current-limiting factor in forward bias.

In order to model the conductance of the device, several factors must be considered regarding the GSDs that have been developed. One of the primary factors that must be considered is the user-defined parameter, $g_c$. This parameter is responsible for describing the overall shape of a device's conductance curve with respect to time. All devices studied in [12]-[18] normally either possess either an inverse exponential, linear, or sigmoidal curve shape (or somewhere between). This curve shape can be influenced by other things than $g_c$ (as shown in Fig. 2), but $g_c$ acts as the primary fitting parameter in order to ensure the device potentiates as expected. Most devices lie in the spectrum between inverse exponential and linear [12]-[18]. A select few devices lie between linear and sigmoidal [13], [18], but they can appear as exponential curve-fits. This model assumes that the conductance of the device has a limit and must saturate at some point, and therefore uses a sigmoidal curve instead of exponential. The assumption being made is that if an exponential conductance curve is being seen, that is the bottom half of a sigmoidal curve, and that if the device were to be further potentiated, its conductive state would saturate with time.

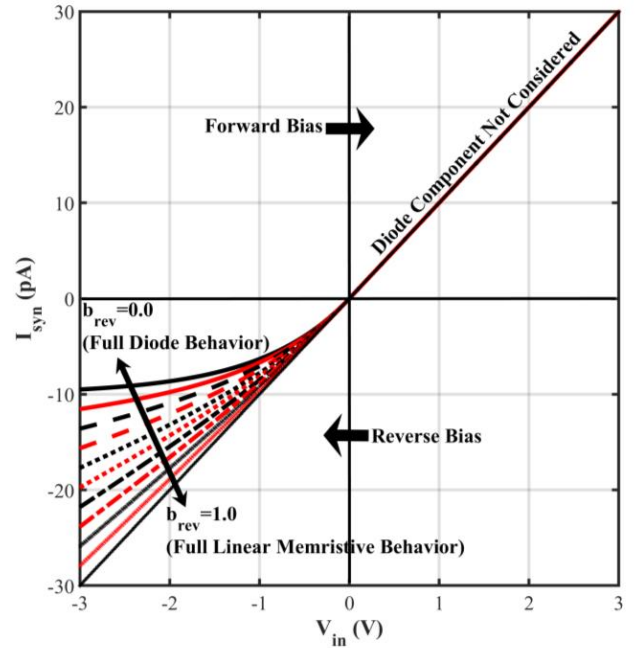

Figure 3. IV curve of an unpotentiated GSD ($g_{syn}$=1e-11S) showing how the term $b_{rev}$ manipulates the current behavior during reverse bias. The device can either exhibit diode-like or linear-like behavior (along with a range of intermediate behavior) while under reverse bias. The $b_{rev}$ term is not considered when the device is in forward bias due to the diode component of the device channel not being the limiting factor in current while in the forward bias state.

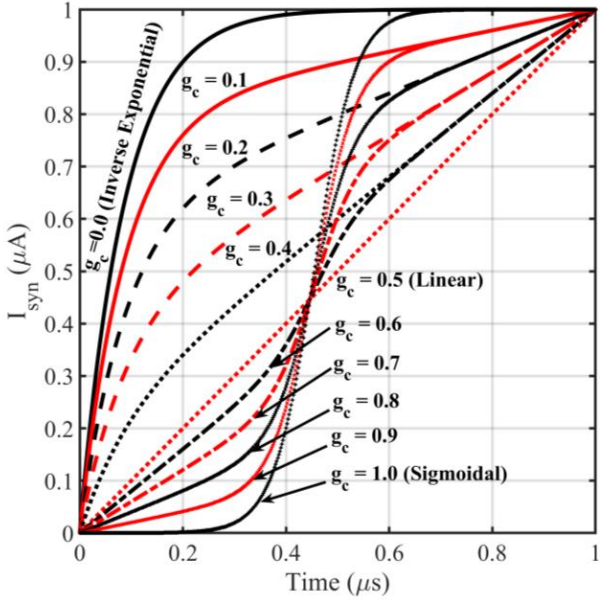

Figure 4. Figure showing how $g_c$ changes the general shape of a conductance curve during constant potentiation over time ($v_{gate}$=1V) where $v_{in}$=1V. Most devices lie in the inverse exponential to linear range ($0 \leq g_c \leq 0.5$), while a select few lie in the linear to sigmoidal range ($0.5 \leq g_c \leq 1.0$).

To model this spectrum of conductance curves, $g_{syn}$ is governed by the following equation

$$g_{syn} = \max(1 - 2g_c, 0)\left(g_{range}(1 - e^{-px})\right) \\ + (-abs(2g_c - 1) + 1)(g_{range}x + g_{min}) \\ + \max(2g_c - 1, 0)\frac{g_{max}}{1 + e^{-mx+s}} \quad (3).$$

Within this equation, the two *max* functions and the absolute value component create a linear combination spectrum to guide the general shape of the conductance curve with respect to time (Fig. 4). The other three components are responsible for getting the conductance from its minimum to maximum value in the desired fashion (inverse exponential, linear, or sigmoidal). These components are all guided by a single unitless state variable, $x$, that can range in value from 0 (initial value for $x_{min}$) and 1 (value for $x_{max}$).

### B. Normalization Constants

In order to universally bound $g_{min}$ and $g_{max}$ to $x$'s 0 to 1 range, the inverse exponential and sigmoidal components of (3) must include pre-calculated normalization terms. These terms within (3) are $s$, $m$, and $p$. The first of these terms, $s$, is calculated by

$$s = \ln(g_{max}/g_{min} - 1) \quad (4).$$

The term, $s$, is used both within the sigmoidal component of (3) and the calculation of the other sigmoidal normalization constant, $m$, which is given by

$$m = \ln(1/g_{range} - 1) + s \quad (5).$$

Within (5) the $g_{range}$ term is simply the difference between the maximum and minimum conductance values,

$$g_{range} = g_{max} - g_{min} \quad (6).$$

Finally, in order to properly normalize the inverse exponential component of (3), $p$ is introduced which is given by

$$p = -\ln(g_{min}/g_{range}) \quad (7).$$

### C. Controlling the State Variable

The unitless state variable, $x$, is responsible for creating the model's memristive behavior. In order to first determine the scale at which $x$ can change between two timesteps, a term named $x_{scale}$ is calculated by

$$x_{scale} = \Delta t/t_{set} \quad (8),$$

where $\Delta t$ is simply the difference in time between the current and previous time stamps within the transient simulation where the model is evaluated. Another component key to calculating $x$ is determining the effective voltage applied to the model's gate terminal. This voltage is defined as $V_{eff}$ and given by

$$V_{eff} = fv_{gate} - o_c\Delta V \quad (9).$$

As a reminder the parameter, $f$, is often 1, and therefore can be largely ignored. The only exception to this rule would be if the device being modeled exhibits behavior where it requires inverted potential applied to its gate in order to potentiate/depress it (e.g. Burgt et al. [13]). The other user-defined parameter within this equation, $o_c$, can be used to model devices such as [16], [17] where potentials applied to different ends of the primary device channel can influence the conductive state of the device. If both $f$ and $o_c$ are kept at their default values (1 and 0, respectively), (9) can be read as $V_{eff}=v_{gate}$.

If the absolute value calculated from (9) is greater than the user-defined threshold voltage for the device ($v_t$), an update to the state variable, $x$, can occur. Prior to the update to $x$ (defined as $\Delta x$), the polarity of $V_{eff}$ is first checked. If $V_{eff}<0$V, the user-defined parameter $n_{amp}$ is applied to $V_{eff}$.

$$V_{eff} = n_{amp}V_{eff} \ (if \ V_{eff} < 0 \ and \ abs(V_{eff}) > v_t) \quad (10).$$

At $n_{amp}$'s default value ($n_{amp}=1$), this equation changes nothing about $V_{eff}$'s value. If $n_{amp}>1$ however, it will exaggerate the negative bias applied to the device in order to more quickly reset the device (e.g. [15]-[17]).

At the start of each simulation, $x$ begins as 0 (unless otherwise initialized by the user-defined parameter, $x_{start}$). Whenever the model is evaluated and $abs(V_{eff})>v_t$, $\Delta x$ is calculated by the equation

$$\Delta x = x_{scale}(V_{eff} - sign(V_{eff})t_cv_t) \quad (11).$$

The *sign* function used in (11) exists in order to ensure $x$ is increased/decreased correctly when the proper potential is applied to the device when $v_t>0$V. The parameter, $t_c$, can be used to tune the defined $v_t$'s effect on $\Delta x$ and determine if it is a hard- or soft-threshold value.

### D. Short- and Long-Term Plasticity

As defined in both Section II and Table I, the model must include forms of both short- and long-term plasticity in order to properly model a GSD. Within this model, short-term plasticity is modeled as a decay term on $x$ at each timestep while long-term plasticity is modeled as change in the minimum value $x$ can obtain ($x_{min}$) at any given time.

Short-term plasticity is created within the model by introducing a decay term, $d_{stp}$, that is taken from $x$ every time the model is evaluated. This decay term is state based in nature, and is given by the equation

$$d_{stp} = r_{stp} t_{set} (x - x_{min}) \Delta t \quad (12).$$

In order to implement long-term plasticity into the model, a term that has so far remained at 0, $x_{min}$, must be given capability to be changed. In devices such as [16], [17], long-term plasticity does exist, but this plasticity can be quickly erased via a depression or reset signal applied to the device due to de-trapping. This evidence shows that long-term plasticity can be potentiated/depressed in a very similar fashion to the primary state variable, $x$. Therefore, the change in $x_{min}$ ($\Delta x_{min}$) can be executed via a very similar equation to (11) under the same condition that $abs(V_{eff}) > v_t$ in order to calculate $\Delta x_{min}$.

$$\Delta x_{min} = q_{ltp} x_{scale} (V_{eff} - sign(v_{eff}) t_c v_t) \quad (13).$$

The only difference in the setup of (13) to (12) is that $\Delta x_{min}$ possesses the user-defined parameter, $q_{ltp}$. At $q_{ltp}$'s default value of 0, no change in $x_{min}$ will ever occur. When $q_{ltp}$ is given a value greater than zero, long-term plasticity will begin to appear within the device over time at a rate dictated by $q_{ltp}$.

Just like $x$, $x_{min}$ can experience decay over time. This can be seen in the long-term plasticity demonstrated in devices such as [11], [16], [17] where the device appears to eventually settle to a final state, but still decays very slowly over time. This decay, $d_{ltp}$, is always removed from $x_{min}$ at every simulation timestep (just like $d_{stp}$) and is defined by

$$d_{ltp} = r_{ltp} t_{set} \Delta t \quad (14).$$

The decay of $x_{min}$ defined in (14) is modeled as a linear decay instead of the state-based decay that is used for $d_{stp}$. This linear modeling of $d_{ltp}$ is primarily due to the often-low magnitude long-term decay normally possesses along with the lack of study on how long-term decay operates in the devices studied [12]-[18]. Linear modeling of $d_{ltp}$ for the work done here proved sufficient in modeling any case where long-term decay appeared.

## IV. FITTING MODEL TO EXPERIMENTAL DATA

In order to fully encapsulate the model described in Section III, a Verilog-A module was developed. The structure and flow of this module can be seen in Algorithm 1. The behavioral block portion of module is the section that is executed at each time step of the simulation, while everything prior to the behavioral block is initialized at the start of the simulation.

The module within Algorithm 1 was used to replicate 26 experimental curves reported in [12]-[18] in order to determine if the model described in this work is robust enough to properly model the types of devices mentioned within each previously published study. The experiments chosen are of various forms including potentiation tests, potentiation/depression curves, potentiation/decay curves, potentiation/decay/reset curves, IV-sweeps on the gate terminal, and IV-sweeps on the input terminal. The results of these 26 replicated experiments can be seen in Fig. 5, while the user-defined parameters used for the module to achieve each test replication can be found in Table II.

## V. RESULTS AND DISCUSSION

While each experimentally reported curve was being replicated in Fig. 5 using the proposed model, the device being presented within each individual publication was assumed to be a single type of device unless otherwise specified. Ideally under this assumption, one would expect to use the same set of model parameters to replicate all experimentally reported characteristics pertaining to a specific publication. As can be seen in Table II, this is not always the case due to several reasons. The first of these reasons is that the study focuses on devices of different sizes or designs such as in [18]. Other reasons for these variances include potential measurement or experiment-specific factors not mentioned in the specific publication or minor secondary effects occurring within a device such as non-uniform transport that the model doesn't capture.

In Figs. 5a-d, work from Bao et al. [12] is replicated. The device used in this study uses an unconventional device channel in the form of a liquid electrolyte. The device also has an extra body terminal which is assumed to be grounded/ignored in the replication simulations performed in this work. The model can exhibit similar behavior to the original experiments when a threshold voltage of $v_t$=0.7V is introduced into the device. Though Bao et al.'s work did not mention a specific threshold voltage for its gate terminal, including the threshold voltage proved key in obtaining similar behavior within Fig. 5a. Long term potentiation was also achieved in Fig. 5c thanks to $q_{ltp}$.

Figs. 5e-i show five different experiments performed by Burgt et al. [13] on a device that requires inverted potential applied to its gate (i.e. $f$=-1). Fig. 5e shows an experiment where the device starts in a semi-potentiated state and then can be potentiated/depressed between five different conductance levels. The remainder of the experiments are typical potentiation/depression or potentiation/decay curves. The one experiment that should be mentioned of these four is the experiment conducted in Fig. 5g. In the original experiment, it is stated that the stimulus provided to the gate in this specific experiment is current, not voltage. All other experiments

| **Algorithm 1** GSD_module ($v_{in}$, $v_{out}$, $v_{gate}$) |
|---|
| 1: **Define** User-Defined Parameters ($v_t$, $b_{rev}$, $g_{min}$, $g_{max}$, $t_{set}$, $r_{stp}$, $g_c$, $n_{amp}$, $o_c$, $t_c$, $q_{ltp}$, $r_{ltp}$, $x_{start}$, $f$) |
| 2: **Define/Calculate** Other Simulation Parameters ($g_{range}$, s, m, p, $x_{max}$) |
| 3: **Define** Variables ($v_{eff}$, x, $g_{syn}$, $t_{curr}$, $t_{past}$, $x_{scale}$, $x_{min}$) |
| 4: **Begin Behavioral Block** |
| 5:     **Calculate** $x_{scale}$ and $v_{eff}$ |
| 6:         abs($v_{eff}$)>$v_t$? |
| 7:             $v_{eff}$<0? |
| 8:                 **Apply** h to $v_{eff}$ |
| 9:         **Calculate** x and $x_{min}$ |
| 10:     x>$x_{min}$? |
| 11:         **Calculate** $d_{stp}$ and **Apply** to x |
| 12:     $x_{min}$>0? |
| 13:         **Calculate** $d_{ltp}$ and **Apply** to $x_{min}$ |
| 14:     **Bound** $x_{min}$ between 0 and $x_{max}$ |
| 15:     **Bound** x between $x_{min}$ and $x_{max}$ |
| 16:     **Calculate** $g_{syn}$ |
| 17:     $v_{in}$-$v_{out}$≥0? |
| 18:         **Calculate** $I_{syn}$ (do not include diode component) |
| 19:     else |
| 20:         **Calculate** $I_{syn}$ (include diode component) |
| 21: **End Behavioral Block** |

TABLE II
USER-DEFINED PARAMETERS FOR EXPERIMENT REPLICATIONS (FIG. 5)

| Paper | Figure | Fig. 5 Subfigure | $g_c$ | $v_t$ | $b_{rev}$ | $g_{min}$ | $g_{max}$ | $t_{set}$ | $r_{stp}$ | $n_{amp}$ | $o_c$ | $t_c$ | $q_{ltp}$ | $r_{ltp}$ | $f$ | $x_{start}$ |
|---|---|---|---|---|---|---|---|---|---|---|---|---|---|---|---|---|
| Bao et al. [12] | 2c | a | 0.40 | 0.700 | 1 | 3.000e-11 | 2.10e-6 | 1800 | 3.5e-3 | 1 | 0.0 | 1 | 0.040 | 7.0e-9 | 1 | 0.0 |
| | 3a | b | 0.40 | 0.700 | 1 | 9.000e-10 | 2.60e-6 | 100 | 3.5e-3 | 1 | 0.0 | 1 | 2.5e-3 | 7.0e-8 | 1 | 0.0 |
| | 3b | c | 0.40 | 0.700 | 1 | 7.000e-10 | 2.60e-6 | 100 | 3.5e-3 | 1 | 0.0 | 1 | 2.5e-3 | 7.0e-8 | 1 | 0.0 |
| | 3d | d | 0.40 | 0.700 | 1 | 3.000e-11 | 1.00e-7 | 4600 | 2.0e-6 | 1 | 0.0 | 1 | 2.5e-3 | 7.0e-8 | 1 | 0.0 |
| Burgt et al. [13] | 1d | e | 0.00 | 0.000 | 1 | 5.750e-4 | 1.35e-3 | 4 | 2.0e-3 | 1 | 0.0 | 0 | 0.400 | 1.0e-6 | -1 | 0.2 |
| | 2a | f | 0.00 | 0.000 | 1 | 5.250e-4 | 1.60e-3 | 1 | 2.0e-3 | 1 | 0.0 | 0 | 0.400 | 1.0e-6 | -1 | 0.0 |
| | 2b | g | 0.60 | 0.000 | 1 | 7.500e-4 | 3.00e-3 | 33 | 1.0e-4 | 1 | 0.0 | 0 | 0.400 | 1.0e-6 | -1 | 0.0 |
| | 2c | h | 0.00 | 0.000 | 1 | 5.250e-4 | 1.60e-3 | 1 | 3.0e-2 | 1 | 0.0 | 0 | 0.100 | 1.0e-7 | -1 | 0.0 |
| | 3c | i | 0.00 | 0.000 | 1 | 1.725e-3 | 7.00e-3 | 1 | 1.0e-2 | 1 | 0.0 | 0 | 0.400 | 1.0e-6 | -1 | 0.0 |
| Herrmann et al. [14] | 2a | j | 0.45 | 0.788 | 1 | 6.000e-12 | 6.00e-9 | 90 | 2.0e-2 | 1 | 0.0 | 1 | 0.010 | 7.0e-8 | 1 | 0.0 |
| | 2b | k | 0.45 | 0.788 | 1 | 6.000e-12 | 6.00e-9 | 90 | 2.0e-2 | 1 | 0.0 | 1 | 0.010 | 1.0e-8 | 1 | 0.0 |
| | 3 | l | 0.45 | 0.788 | 1 | 6.000e-12 | 6.00e-9 | 90 | 1.0e-3 | 1 | 0.0 | 1 | 0.010 | 1.7e-6 | 1 | 0.0 |
| | 4a | m | 0.45 | 0.788 | 1 | 6.000e-12 | 6.00e-9 | 90 | 2.0e-2 | 1 | 0.0 | 1 | 0.010 | 1.7e-6 | 1 | 0.0 |
| Lim et al. [15] | 2a | n | 0.45 | 0.000 | 0 | 1.000e-12 | 2.00e-9 | 5.5e-3 | 1.0e-1 | 40 | 0.0 | 0 | 0.000 | 0.0 | 1 | 0.0 |
| | 2b | o | 0.45 | 0.000 | 0 | 1.000e-12 | 2.00e-9 | 5.5e-3 | 1.0e-1 | 40 | 0.0 | 0 | 0.000 | 0.0 | 1 | 0.0 |
| | 5 | p | 0.45 | 0.000 | 0 | 1.000e-12 | 2.00e-9 | 5.5e-3 | 1.0e-1 | 40 | 0.0 | 0 | 0.000 | 0.0 | 1 | 0.0 |
| Murdoch et al. [16] | 4c | q | 0.05 | 2.000 | 0 | 2.500e-10 | 1.40e-9 | 5 | 1.2e-1 | 1 | 1.0 | 1 | 0.020 | 3.0e-4 | 1 | 0.0 |
| | 4d | r | 0.05 | 1.990 | 0 | 3.000e-9 | 1.15e-8 | 3 | 4.5e-1 | 40 | 1.0 | 1 | 0.040 | 3.0e-4 | 1 | 0.0 |
| | 5a | s | 0.05 | 2.000 | 0 | 2.800e-9 | 1.00e-8 | 3 | 8.5e-1 | 40 | 1.0 | 1 | 0.100 | 7.0e-3 | 1 | 0.0 |
| Tan et al. [17] | 1c | t | 0.05 | 1.400 | 1 | 5.000e-12 | 4.00e-8 | 5500 | 1.0e-8 | 345 | 0.0 | 1 | 0.010 | 1.0e-6 | 1 | 0.0 |
| | 1d | u | 0.05 | 1.400 | 1 | 5.000e-12 | 4.00e-8 | 2500 | 6.0e-4 | 345 | 1.0 | 1 | 0.175 | 2.0e-8 | 1 | 0.0 |
| | 1e | v | 0.00 | 0.800 | 1 | 1.000e-13 | 4.00e-8 | 5500 | 7.0e-5 | 345 | 0.5 | 1 | 0.250 | 1e-10 | 1 | 0.0 |
| Tang et al. [18] | 2 | w | 0.85 | 0.000 | 1 | 1.000e-9 | 2.40e-9 | 1175 | 2.0e-7 | 1 | 0.0 | 0 | 0.000 | 0.0 | 1 | 0.0 |
| | 9 | x | 0.00 | 0.000 | 1 | 2.040e-9 | 4.50e-9 | 50 | 4.0e-5 | 1 | 0.0 | 0 | 0.600 | 1.0e-8 | 1 | 0.0 |
| | 14 | y | 1.00 | 0.000 | 1 | 5.000e-12 | 6.00e-8 | 10 | 9.5e-5 | 1 | 0.0 | 0 | 0.000 | 0.0 | 1 | 0.0 |
| | 15 | z | 1.00 | 0.000 | 1 | 5.000e-11 | 3.00e-9 | 150 | 2.0e-7 | 1 | 0.0 | 0 | 0.000 | 0.0 | 1 | 0.0 |

replicated from Burgt et al.'s work use voltage to program the device. As previously mentioned, the model developed here does not specifically use current bias applied to the gate to program the device, but instead only voltage. Fig. 5g also is the only experiment that uses a $g_c$ value greater than 0.5, meaning it lies in the sigmoidal conductance range. This shows that using current instead of voltage to program the device from Burgt et al. might cause it to exhibit a different form of potentiation at a physics-level, and this change in behavior can be replicated by varying $g_c$.

The next device studied is one developed by Herrmann et al. [14], which can be seen in Figs. 5j-m. The device in Herrmann et al.'s work is a gated resistive-RAM device that uses a device channel made of strontium titanate ($SrTiO_4$) with a defined threshold voltage of $v_t$=0.788V [14]. The experiments conducted in this work primarily focus on sweeping or potentiating voltages rather than pulse testing. Extremely similar results to Herrmann et al.'s original work were obtained in Figs. 5j and 5l. Similar results were also obtained during the experiments in Figs. 5k and 5m, although a couple small behaviors were not successfully captured from these two experiments. In Fig. 5k, the original experiment showed plateauing of the synaptic current as it reached higher values of $v_{in}$. This could be accomplished via the GSD model, but not without severely compromising the current values while $v_{in}$<0V. One potential issue with the original experiment that Fig. 5k replicates is the programming time and length of the device are not specifically mentioned for the gate voltages shown. The original experiment Fig. 5m replicates also saw very slight decay prior to plateauing at the lower gate voltages of $v_{gate}$=1V and 2V, which was not able to be replicated with the GSD model.

Within Figs. 5n-p experiments replicated from Lim et al. [15] can be seen. The device uses a set of gated diodes within CMOS in order to create memristive behavior. The device possesses highly linear potentiation/depression behavior while in forward bias (Figs. 5o and 5p), albeit with a high $n_{amp}$ value ($n_{amp}$=40) that creates a very asymmetric potentiation/depression curve. The device also experiences typical diode behavior while under reverse bias conditions (Fig. 5n) since diodes are core to the device's functionality.

The experiments modeled in Figs. 5q-s are from a device developed by Murdoch et al. [16] and is the first of two light-gated synaptic devices studied in this work. Instead of light to potentiate the device, the GSD model simply uses a substitute voltage of 5V as a gate programming voltage. The experiments for Murdoch et al.'s device primarily revolve around potentiate/decay tests with the experiment in Fig. 5r including a reset signal that is applied on the device's output node. This behavior means that this device has an $o_c$ value of 1. The Murdoch et al. device also had a low $t_{set}$ value with respect to some of the other devices studied but possessed a very low ON/OFF ratio between its $g_{min}$ and $g_{max}$ values.

Figs. 5t-v show the second light-gated device that was studied in this model's development. Developed by Tan et al. [17], this light-gated device also possesses a low ON/OFF ratio between the $g_{min}$ and $g_{max}$ values for the device (similar to Murdoch et al.'s device), but with a much higher $t_{set}$ value. During experiment replication, a 5V gate bias is used once again to imitate light stimulation to the device. In Fig. 5t, a range of voltages were used to mimic the same range of light intensities used in the original experiment. The range of voltages follows the same ratio the light intensities follow with respect to the highest light intensity used. Like Murdoch et al.'s

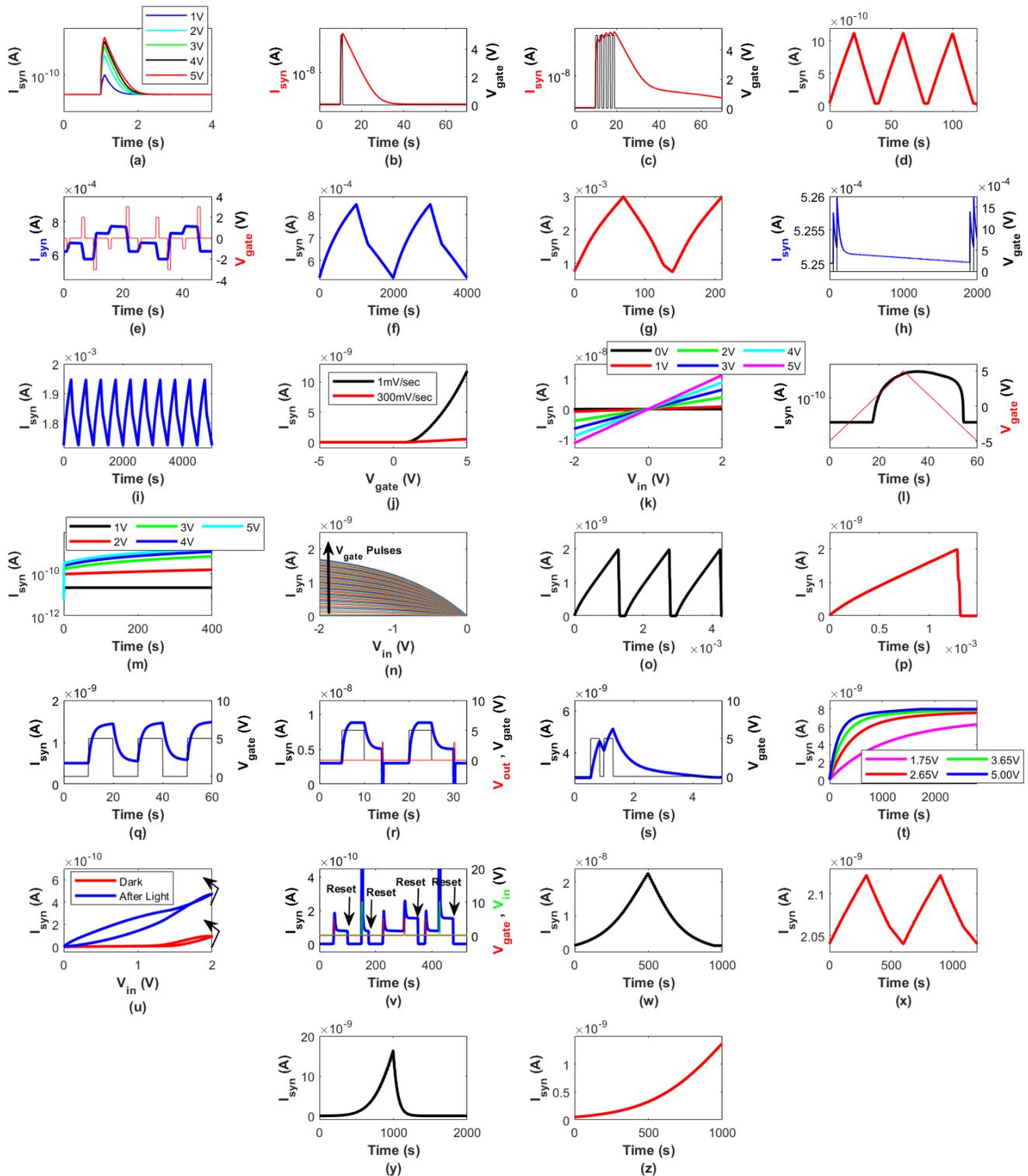

Figure 5. Replicated experiments from previously published papers. (a) Fig. 2c from [12] where bias is applied to the gate for 100ms when t=1sec at five different gate voltages. (b) Fig. 3a from [12]. (c) Fig. 3b from [12]. (d) Fig. 3d from [12] showing three consecutive highly linear potentiation/depression curves. (e) Fig. 1d from [13]. (f) Fig. 2a from [13] showing two consecutive potentiation/depression curves. (g) Fig. 2b from [13] showing 1.5 potentiation/depression curves. (h) Fig. 2c (inset) from [13]. (i) Fig. 3c from [13] showing ten consecutive potentiation/depression curves. (j) Fig. 2a from [14] where the gate voltage is swept at two different velocities. (k) Fig. 2b from [14] where the input voltage is swept after the device has been pre-programmed for a fixed period (2000s) with six different gate voltages. (l) Fig. 3 from [14]. (m) Fig. 4a from [14] where the device was potentiated over time with five different constant gate voltages. (n) Fig. 2a from [15] where the device undergoes a reverse bias sweep on its input node between 64 separate programming pulses to its gate. (o) Fig. 2b from [15] showing three consecutive potentiation/depression curves. (p) Fig. 5 from [15] showing one potentiation/depression curve. (q) Fig. 4c from [16]. (r) Fig. 4d from [16] where the device is reset with its output node when t=14sec and t=30sec. (s) Fig. 5a from [16]. (t) Fig. 1c from [17] where the device is potentiated by four different constant gate voltages over time. (u) Fig. 1d from [17] showing an IV curve before and after "light" (i.e. a voltage pulse in the replication) is applied to the gate. (v) Fig. 1e from [17] where the device is reset using its input terminal at the four points specified. (w) Fig. 2 from [18] showing one potentiation/depression curve. (x) Fig. 9 from [18] showing two potentiation/depression curves. (y) Fig. 14 from [18] showing one potentiation/depression curve. (z) Fig. 15 from [18] showing one potentiation curve using voltage pulses.

device, this device can be reset via a bias applied on the primary device channel. This behavior means that a non-zero value for $o_c$ was needed during tests where resets occur (such as Fig. 5v). In Fig. 5u, a pair of IV curves are shown before and after a pulse of "light" is exposed to the device (or in the replication a voltage pulse to the gate). In the original experiment (Fig. 1d in [17]), a phenomenon known as *negative differential resistance* (NDR) can be seen where conductance of the device continues to increase despite the voltage applied to the input node decreasing. The GSD model in Fig. 5u was only able to replicate this phenomenon to a very minor degree, and not to the level seen in the original experiment. This failure of replicating NDR is primarily due to how decay of both short- and long-term plasticity works within the GSD model.

The final device studied and shown in Fig. 5 is the one developed by Tang et al. [18] shown in subfigures w-z. This device, dubbed an "electro-chemical RAM" (ECRAM) device, uses a simple device gate for potentiation/depression. Most experiments on the ECRAM device focus on current bias programming pulses instead of a voltage bias. Within Tang et al.'s work, multiple different sized devices were also studied, with some showing different behavior than others. Within the experiments replicated for Tang et al., voltage pulses were used at similar ratio levels with respect to the ratio levels of the current pulses used in the original experiments. The maximum current pulse magnitude in the original experiments was 1mA, and the maximum voltage amplitude in the replicated version of each experiment was 1V (i.e. if an original experiment used current pulses of 1μA, the replicated experiment would use voltage pulses at 1/1000$^{th}$ the magnitude of the highest voltage used, 1V). Some of the ECRAM devices shown in [18] possess more sigmoidal conductance curve behavior (Figs. 5w, y, and z where $0.5 < g_c \leq 1.0$), while others possess inverse exponential behavior (Fig. 5x where $0.0 \leq g_c < 0.5$). Device sizing on the ECRAM device could be responsible for these different behaviors seen but cannot be stated for certain due to the original work not stating the size of each device studied in every figure.

With all seven devices studied, the GSD model can encapsulate most critical elements of each device's behavior with a few exceptions. Some of the plateauing seen in Herrmann et al.'s device [14] was not able to be encapsulated (Fig. 5k) along with negative-differential resistance phenomenon seen in Tan et al.'s light-gated device [17] (Fig. 5u). Other than those exceptions, the model can robustly replicate GSD behavior in any of the studied devices.

## VI. Conclusion

A comprehensive model for gated-synaptic devices has been proposed and simulated using SPICE/Verilog-A for the first time. Through verification against previously published device experiments, the GSD model shows that it encapsulates critical behaviors such as proper thresholding, short- and long-term plasticity, symmetric and asymmetric potentiation/depression, user-defined conductance ranges, device channel bias influence, and other more specific features. Through this wide range of behavior, the model demonstrates it can provide a great baseline for circuit, architecture, and systems engineers looking to utilize some form of GSD within their design. The model will allow GSDs to be more widely used within neuromorphic hardware that relies on temporal-based behaviors or generic memory architectures that are looking to utilize a different type of non-volatile memory. The ease-of-use of this model can assist in spurring the creation of more neuromorphic architecture designs such as the ROLLS chip [23], IBM's TrueNorth architecture [24], Intel's Loihi chip [25], or Stanford's Braindrop [26]. As neuromorphic hardware becomes easier to develop and create, neural networks can run more efficiently on hardware; reducing the resource overhead that often plagues neural network simulation. Models like the one shown here can help a broad community of researchers in the area of near memory or neuromorphic computing tackle the challenge of jumping into the next decade and start developing more powerful alternatives to Von Neumann architectures.

ACKNOWLEDGMENTS

The authors would like to acknowledge Clare Thiem (Air Force Research Laboratory, Rome, NY) and Cory Merkel (Rochester Institute of Technology, Rochester, NY) for guidance and discussion on this work.

REFERENCES

[1] I. Hubara, M. Courbariaux, D. Soudry, R. El-Yaniv, and Y. Bengio, "Binarized Neural Networks," in *proc. of Advances in Neural Information Processing Systems 29 (NIPS 2016)*, pp. 1-9, Dec. 2016.

[2] A. Graves, A. Mohamed, and G. Hinton, "Speech recognition with deep recurrent neural networks," in *proc. of 2013 IEEE International Conference on Acoustics, Speech and Signal Processing*, pp. 6645-6649, May 2013. doi: ICASSP.2013.6638947.

[3] K. Gurney. *An introduction to neural networks*. CRC press, 2014.

[4] G. P. Zhang, "Avoiding Pitfalls in Neural Network Research," *IEEE Transactions on Systems, Man, and Cybernetics, Part C*, 37 (1), pp. 3-16, Dec. 2006. doi: 10.1109/TSMCC.2006.876059.

[5] C. D. Schuman, T. E. Potok, R. M. Patton, J. D. Birdwell, M. E. Dean, G. S. Rose, and J. S. Plank, "A Survey of Neuromorphic Computing and Neural Networks in Hardware," in *arxiv: 1705.06963*, pp. 1-88, May 2017.

[6] N. Qiao, H. Mostafa, F. Corradi, M. Osswald, F. Stefanini, D. Sumislawska, and G. Indiveri, "A reconfigurable on-line learning spiking neuromorphic processor comprising 256 neurons and 128K synapses," *Front. Neurosci.*, 9 (141), pp. 1-17, Apr. 2015. doi: 10.3389/fnins.2015.00141.

[7] C.-K. Lin, A. Wild, G. N. Chinya, Y. Cao, M. Davies, D. M. Lavery, H. Wang, "Programming Spiking Neural Networks on Intel's Loihi," *Computer*, 51 (3), pp. 52-61, Mar. 2018. doi: 10.1109/MC.2018.157113521.

[8] G. Indiveri, B. Linares-Barranco, R. Legenstein, G. Deligeorgis, and T. Prodromakis, "Integration of nanoscale memristor synapses in neuromorphic computing architectures," *Nanotechnology*, 24 (38), pp. 1-13, Sept. 2013.

[9] D. B. Strukov, G. S. Snider, D. R. Stewart, and R. S. Williams, "The missing memristor found," *nature*, 453, pp. 80-83, May 2008. doi: 10.1038/nature06932.

[10] L. Chua, "Resistance switching memories are memristors," *Appl. Phys. A*, 102 (4), pp. 765-783, Mar. 2011. doi: 10.1007/s00339-011-6264-9.

[11] Z. Wang, S. Joshi, S. E. Savel'ev, H. Jiang, R. Midya, P. Lin, M. Hu, N. Ge, J. P. Strachan, Z. Li, Q. Wu, M. Barnell, G.-L. Li, H. L. Xin, R. S. Williams, Q. Xia, and J. J. Yang, "Memristors with diffusive dynamics as synaptic emulators for neuromorphic computing," *Nature Materials*, 16, pp. 101-108, Sept. 2016. doi: 10.1038/nmat4756.

[12] L. Bao, J. Zhu, Z. Yu, R. Jia, Q. Cai, Z. Wang, L. Xu, Y. Wu, Y. Yang, Y. Cai, and R. Huang, "Dual-Gated $MoS_2$ Neuristor for Neuromorphic Computing," *ACS Appl. Mater. Interfaces*, Oct. 2019, pp. 1-35. doi: 10.1021/acsami.9b10072.

[13] Y. van de Burgt, E. Lubberman, E. J. Fuller, S. T. Keene, G. C. Faria, S. Agarwal, M. J. Marinella, A. A. Talin and A. Salleo, "A non-volatile organic electrochemical device as a low-voltage artificial synapse for neuromorphic computing," *Nature materials*, 16 (4), pp. 414-418, Feb. 2017. doi: 10.1038/nmat4856.

[14] E. Herrmann, A. Rush, T. Bailey, and R. Jha, "Gate Controlled Three-Terminal Metal Oxide Memristor," *IEEE Electron Device Letters*, 39 (4), pp. 500-503, Apr. 2018. doi: 10.1109/LED.2018.2806188.

[15] S. Lim, J.-H. Bae, J.-H. Eum, S. Lee, C.-H. Kim, D. Kwon, and J.-H. Lee, "Hardware-based Neural Networks using a Gated Schottky Diode as a Synapse Device," in *proc. 2018 IEEE International Symposium on Circuits and Systems (ISCAS)*, May 2018, pp. 1-5. doi: 10.1109/ISCAS.2018.8351152.

[16] B. J. Murdoch, T. J. Raeber, Z. C. Zhao, A. J. Barlow, D. R. McKenzie, D. G. McCulloch, and J. G. Partridge, "Light-gated amorphous carbon memristors with indium-free transparent electrodes," *Carbon*, 152, pp. 59-65, Nov. 2019. doi: 10.1016/j.carbon.2019.06.022.

[17] H. Tan, G. Liu, H. Yang, X. Yi, L. Pan, J. Shang, S. Long, M. Liu, Y. Wu, and R.-W. Li, "Light-Gated Memristor with Integrated Logic and Memory Functions," *ACS Nano*, 11, Oct. 2017, pp. 11298-11305. doi: 10.1021/acsnano.7b05762.

[18] J. Tang, D. Bishop, S. Kim, M. Copel, T. Gokmen, T. Todorov, S. Shin, K.-T. Lee, P. Solomon, K. Chan, W. Haensch and J. Rozen, "ECRAM as Scable Synaptic Cell for High-Speed, Low-Power Neuromorphic Computing," in *2018 IEEE International Electron Devices Meeting (IEDM)*, pp. 13.1.1-13.1.4, Dec. 2018. doi: 10.1109/IEDM.2018.8614551.

[19] R. Tyasnurita, E. Özcan and R. John, "Learning Heuristic Selection using a Time Delay Neural Network for Open Vehicle Routing," in *2017 IEEE Congress on Evolution Computation (CEC)*, pp. 1474-1481, Jul. 2017, San Sebastian, Spain. doi: 10.1109/CEC.2017.7969477.

[20] M. Shahsavari and P. Boulet, "Parameter Exploration to Improve Performance of Memristor-Based Neuromorphic Architectures," *IEEE Transactions on Multi-Scale Computing Systems*, 4 (4), pp. 833-846, Oct. 2017. doi: 10.1109/TMSCS.2017.2761231.

[21] M. Boerlin and S. Denève, "Spike-Based Population Coding and Working Memory," *PLoS Comput Biol*, 7 (2), Feb. 2011. doi: 10.1371/journal.pcbi.1001080.

[22] L. F. Abbott and W. G. Regehr, " Synaptic computation," *Nature*, 431, pp. 796-803, Oct. 2004. doi: 10.1038/nature03010.

[23] G. Indiveri, F. Corradi and N. Qiao, "Neuromorphic architectures for spiking deep neural networks," *2015 IEEE International Electron Devices Meeting (IEDM)*, Washington, DC, 2015, pp. 4.2.1-4.2.4. DOI: 10.1109/IEDM.2015.7409623.

[24] J. Hsu, "Ibm's new brain [news]," *IEEE spectrum*, 51 (10), pp. 17-19, 2014.

[25] C-K. Lin et al., "Programming spiking neural networks on Intel's Loihi," *Computer*, 51 (3), pp. 52-61, 2018.

[26] A. Neckar, S. Fok, B. V. Benjamin, T. C. Stewart, N. N. Oza, A. R. Voelker, C. Eliasmith, R. Manohar and K. Boahen, "Braindrop: A Mixed-Signal Neuromorphic Architecture With a Dynamical Systems-Based Programming Model," in *Proceedings of the IEEE*, 107 (1), pp. 144-164, Jan. 2019. doi: 10.1109/JPROC.2018.2881432.